\documentclass[11pt,a4paper]{article}
\usepackage[hyperref]{acl2020}
\usepackage{times}
\usepackage{latexsym}

\usepackage{microtype}

\usepackage{makecell}
\usepackage{todonotes}
\usepackage{xcolor}
\usepackage{tikz, pgfplots}
\usepackage{graphicx}
\usepackage{booktabs}
\usepackage{amssymb}
\usepackage{amsmath}
\usepackage{multirow}
\usepackage{float}

\usepackage{pgf}
\usetikzlibrary{arrows,shapes.misc,chains,scopes}
\usepackage{pgfplotstable}
\pgfplotsset{compat=1.15}

\aclfinalcopy 

\title{Context Matters in Semantically Controlled Language Generation for Task-oriented Dialogue Systems}

\author{Ye Liu$^1$, Wolfgang Maier$^2$, Wolfgang Minker$^3$ \and Stefan Ultes$^4$ \\
 $^{1,2,4}$Mercedes-Benz AG, Sindelfingen, Germany \\
 $^{1,3}$Ulm University, Ulm, Germany \\
 \texttt{ $^1$ye.y.liu@daimler.com} \\
 \texttt{ $^{2,4}$\{wolfgang.mw.maier,stefan.ultes\}@daimler.com}\\
 \texttt{ $^3$wolfgang.minker@uni-ulm.de}}
 
\date{}

\begin{document}

\maketitle
\begin{abstract}
This work combines information about the dialogue history encoded by pre-trained models with a meaning representation of the current system utterance to realize contextual language generation in task-oriented dialogues. We utilize the pre-trained multi-context ConveRT model for context representation in a model trained from scratch; and leverage the immediate preceding user utterance for context generation in a model adapted from the pre-trained GPT-2. Both experiments with the MultiWOZ dataset show that contextual information encoded by pre-trained models improves the performance of response generation both in automatic metrics and human evaluation. Our presented contextual generator enables higher variety of generated responses that fit better to the ongoing dialogue. Analysing the context size shows that longer context does not automatically lead to better performance, but the immediate preceding user utterance plays an essential role for contextual generation. In addition, we also propose a re-ranker for the GPT-based generation model. The experiments show that the response selected by the re-ranker has a significant improvement on automatic metrics.
\end{abstract}

\section{Introduction}
In a conversation, speakers are influenced by previous utterances and tend to adapt their way of speaking to each other \citep{duvsek2016context, reitter2006computational}. Furthermore, generating the responses that fit well to dialogue context facilitates successful conversation and strengthens the user's impression of Spoken Dialogue Systems (SDSs). Several previous works \citep{duvsek2016context, kale2020template, sankar2019neural} have explored the impact of previous dialogue information on the generated language in task-oriented dialogue. However, how to efficiently infuse the dialogue context into a semantically controlled generator for improving the contextual interactive experience is still challenging. Such as, the contextual generator proposed in \citet{duvsek2016context} has no big improvement without the help of an additional re-ranker. The empirical study in \citet{sankar2019neural} demonstrated that both recurrent and transformer-based seq2seq model can not effectively consider previous dialogue history for generation. In this work, we propose two contextual generators, which both utilize pre-trained models to encode dialogue context. And the experiment results show that context does matter in semantically controlled task-oriented Natural Language Generation (NLG).

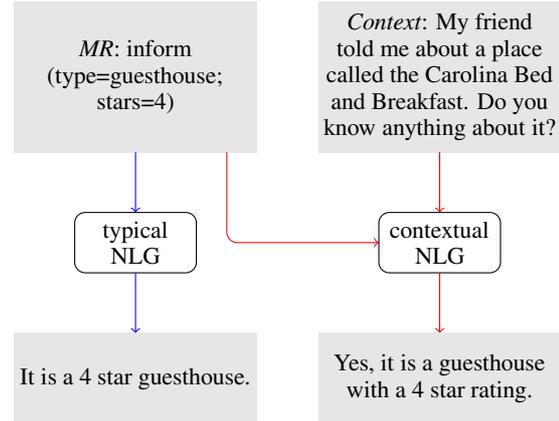
\begin{figure}
\centering
\begin{tikzpicture}[scale=0.8]
\footnotesize
\fill [fill=gray!20] (-4.5,-1) rectangle (-0.5,1.5);
\node at (-2.5,0.25)[align=center]  
{\emph{MR}: inform\\(type=guesthouse; \\ stars=4)};
\fill [fill=gray!20] (0.5,-1) rectangle (4.5,1.5);
\node at (2.5,0.25)[align=center]  
{\emph{Context}: My friend\\ told me about a place\\ called the Carolina Bed\\ and Breakfast. Do you\\ know anything about it?};
\draw[blue, ->] (-2.5,-1) -- (-2.5,-2);
\draw[rounded corners]  (-3.5,-3) rectangle (-1.5,-2);
\node at (-2.5,-2.5)[align=center]  {typical\\NLG};
\draw[red, ->] (2.5,-1) -- (2.5,-2);
\draw[red, ->,rounded corners] (-1,-1) -- (-1,-2.5) -- (1.5,-2.5);
\draw[rounded corners]  (1.5,-3) rectangle (3.5,-2);
\node at (2.5,-2.5)[align=center]  {contextual\\NLG};
\draw[blue, ->] (-2.5,-3) -- (-2.5,-4);
\draw[red, ->] (2.5,-3) -- (2.5,-4);
\fill [fill=gray!20] (-4.5,-5.5) rectangle (-0.5,-4);
\node at (-2.5,-4.75)[align=center]  
{It is a 4 star guesthouse.};
\fill [fill=gray!20] (0.5,-5.5) rectangle (4.5,-4);
\node at (2.5,-4.75)[align=center]  
{Yes, it is a guesthouse \\ with a 4 star rating.};
\end{tikzpicture}
    \caption{An example of contextual NLG (red part) compared with typical NLG (blue part) in our experiments. (More examples with multi-turn context please find in the Appendix)}
    \label{fig:an example of contextual NLG compared with typical NLG}
\end{figure}

The function of NLG in task-oriented SDS is to generate meaningful output in the form of natural language with the guidance of meaning representation (MR). The MR is a formalism of response semantics and generally represents a dialogue action (DA), such as \emph{inform} or \emph{request}, along with one or more slots and their values (See the MR in Figure \ref{fig:an example of contextual NLG compared with typical NLG}). However, this typical NLG only takes the input MR into account and has no clue on how to adapt to the dialogue history. This results in coarse and flat responses (see the blue part in Figure \ref{fig:an example of contextual NLG compared with typical NLG}). To enable the contextual interaction in SDS, task-oriented NLG should not only contain the desired MR information, but also have access to the dialogue history. The example shown in Figure \ref{fig:an example of contextual NLG compared with typical NLG} taken from our experiments: compared with non-contextual generation, the response with context guidance adapts better to the preceding dialogue and is more like a sentence from a real human. In addition, the contextual NLG models are prone to generate more diverse responses.

In summary, the main contributions of this paper are as follows:
\begin{itemize}
  \item
  To leverage the contextual nature of the multi-turn dialogue, we utilize the pre-trained multi-context ConveRT~\citep{henderson2020convert} to encode dialogue history for the contextual generator. These contextual embeddings are then forwarded to the \textbf{S}emantically \textbf{C}ontrolled \textbf{LSTM} (SC-LSTM) generator~\citep{wen2015semantically}. The \textbf{C}onveRT initialized \textbf{SC-LSTM} is called \emph{CSC-LSTM} for short in this paper. With the powerful multi-context encoding of ConveRT, we also analyse the impact of variable context size on \emph{CSC-LSTM}. To the best of our knowledge, we are the first to utilize the pre-trained conversational model ConveRT for contextual generation in task-oriented dialogue system.
  
  \item
  We leverage only the immediate preceding user utterance for contextual generation. Adapted from GPT-2~\citep{radford2019language}, the user utterance and DA guide the contextual generation as context and semantic information respectively. We call \textbf{C}ontextually and \textbf{S}emantically \textbf{C}onditioned \textbf{GPT} \emph{CSC-GPT} for short in this paper. The experiments of GPT-based contextual model show that generation benefits from dialogue context, even if only immediate preceding user utterance is taken into account.
 
  \item
  We propose a \textbf{BERT}-based~\citep{devlin2019bert} \textbf{R}e-ranker (\emph{BERT-R}) for the \emph{CSC-GPT} generator, to select system response that fit better to the user utterance. Given the top $5$ generations of \emph{CSC-GPT}, several automatic scores are regarded as contextual basis between user utterance and system response to train on a multiple regression task adapted from BERT. Experiments show that the re-selected generation has a significant improvement on the performance scores.
\end{itemize}

The remainder of this paper is structured as follows: Section~\ref{sec: related worrk} shows the related works of our research. Section~\ref{sec: Dataset and Automatic Metric Scores} introduces the dataset and the automatic metric scores which are used in this work. Section \ref{sec: the proposed models} describes our proposed three models: \emph{CSC-LSTM}, \emph{CSC-GPT} and \emph{BERT-R}, as well as the experiment details. Section \ref{sec: the experiment results} shows the experiment results of all models in automatic metrics and human evaluation. The last Section~\ref{sec:Conclusion and Future Work} concludes and outlines future research.

\section{Related Works}
\label{sec: related worrk}

For task-oriented NLG, semantically controlled neural models play a significant role.~\citet{wen2015semantically} introduced a semantically conditioned model by adding an additional semantic cell in LSTM to control the DA, which is defined as the combination of intent and slot-value pairs, for generation.~\citet{tseng2018variational} improved the RNN-based generator by considering latent information using the semantically conditional variational autoencoder architecture. As the major advantage and superior performance of pre-trained LMs~\citep{devlin2019bert,radford2019language},~\citet{peng2020few} proposed a semantically controlled generation model based on GPT-2;~\citet{chen2020few} and~\citet{peng2020soloist} presented an end-to-end task-oriented SDS based on the pre-trained GPT-2. Even though there has been plenty of works on semantically guided NLG, most approaches fail in utilizing information of the preceding interaction.

\citet{duvsek2016context} stood out as they extended the idea of NLG from MRs by adding \textit{one} preceding user utterance to their recurrent encoder. However, we are more interested in the influence of bigger context sizes in \emph{CSC-LSTM} contextual generation. And the model in \citet{duvsek2016context} was not able to show any improvement for contextual generation without an additional $n$-gram match re-ranker. However, our proposed contextual generators outperform both the baselines even without re-ranker and the re-ranker in our work can further highly improve the generation performance on automatic metrics.~\citet{sankar2019neural} made an empirical study to understand how models use the available dialog history for generation and found that both recurrent and transformer-based seq2seq model can not consider previous dialogue history effectively. However, the dialogue history in our work is encoded by pre-trained models and experiments show that the generation in task-oriented dialogue benefits from dialogue history.~\citet{kale2020template} also examined the role of context and demonstrated that the generation benefits from the dialogue history. While their approach highly relies on manually pre-defined templates which are costly to create, in this work, all responses are directly generated without the need of any templates.

\section{Dataset and Automatic Metrics}
\label{sec: Dataset and Automatic Metric Scores}
In our work, the automatic metric scores are not only used for performance evaluation, but also used for the \emph{BERT-R} training in Section \ref{subsec: BERT re-ranker}. Hence, we introduce the dataset and the automatic metrics beforehand in this section.

\subsection{MultiWOZ dataset}
\label{subsec: MultiWOZ dataset}
The original MultiWOZ~\citep{budzianowski2018multiwoz} dataset is a fully-labeled collection of human-human written conversations spanning over multiple domains and topics. It contains over $10,000$ dialogues spanning $8$ domains, namely: Restaurant, Hotel, Attraction, Taxi, Train, Hospital, Bus, and Police. The test and validation sets contain $1,000$ examples each for performance comparison. The MultiWOZ 2.1~\citep{eric2020multiwoz} and MultiWOZ 2.2~\citep{zang2020multiwoz} both fix some dialogue state annotation errors and dialogue utterances, resulting in an improved version of the original MultiWOZ.

The MultiWOZ 2.1 is used for evaluation of \emph{CSC-LSTM} in Section \ref{subsec: convert initialized sc-ltsm} in accordance with related work. The MultiWOZ 2.1 and 2.2 are both used additionally for evaluating \emph{CSC-GPT} and \emph{BERT-R} in Section \ref{subsec:contextually and semantically Condditioned GPT} and \ref{subsec: BERT re-ranker}.

\subsection{Automatic metrics}
\label{subsec: Automatic metrics}
The following metrics are used for performance comparison in Section \ref{sec: the experiment results} and several are applied for \emph{BERT-R} training as target score in Section \ref{subsec: BERT re-ranker}.

\subsubsection{$\emph{N}$-gram matching metrics} 

\textbf{BLEU-4} is the 4-gram BLEU score~\citep{papineni2002bleu}, which is the most widely used metric score for evaluating the performance of language generation and machine translation. In this work, BLEU-4 is computed for multiple values of $n = (1, 2, 3, 4)$ with weights $(0.25, 0.25, 0.25, 0.25)$ respectively and the scores are averaged geometrically. A smoothing function is used to avoid that no $\emph{n}$-gram overlaps are found. 
    
The target signal $\tau_{\text{BLEU-4}}$ in Section \ref{subsec: BERT re-ranker} represents the BLEU-4 score between system generated response and gold reference in \emph{BERT-R}.

\textbf{Meteor}~\citep{banerjee2005meteor} is a kind of weighted F-score based on mapping unigrams and also computes a penalty function for incorrect word order.~\citet{lavie2007meteor} demonstrated that Meteor score has high correlation with human ratings. 
    
The target signal $\tau_{\text{Meteor}}$ in Section \ref{subsec: BERT re-ranker} represents the Meteor score between system generation and gold reference in \emph{BERT-R}.

\subsubsection{Machine learned metric}

\textbf{BERTScore}~\citep{zhang2019bertscore} is a machine learned automatic evaluation metric for text generation that has shown a high correlation with human judgments. BERTScore leverages the pre-trained contextual embeddings from variants of BERT~\citep{devlin2019bert} and matches words in candidate and reference sentences by cosine similarity. Moreover, BERTScore computes precision, recall, and F1 measure\footnote{Only F1 score, which represents a reasonable balance between recall and precision, is shown in Table~\ref{tab:Results for BLEU-4, Meteor, BERTScore and Variation size of CSC-LSTM model with variant context size.}, Table~\ref{tab:Performance comparison on automatic metrics of all models with top one generation}, Table~\ref{tab:Performance of 0u0s CSC-GPT and 1u1s CSC-GPT.}, Table~\ref{tab:Performance of reslected generations by BERT-R and top one CSC-GPT.} and Figure~\ref{fig:The changeable tendency of all metrics over variable context model.} for performance comparison.}.~\citet{zhang2019bertscore} showed that the Roberta~\citep{liu2019roberta} large model has the best-performing results for English tasks. So the roberta-large model\footnote{https://github.com/Tiiiger/bert\_score} is used for computing BERTScore in this work. 
    
The target signals $\tau_{\text{BERT}_{\text{pre}}}$, $\tau_{\text{BERT}_{\text{rec}}}$, $\tau_{\text{BERT}_{\text{f1}}}$ in Section \ref{subsec: BERT re-ranker} represent the precision, recall and F1 of BERTScore between system response and gold reference respectively in \emph{BERT-R}.

\subsubsection{Other metrics: }

\textbf{{ConveRT cosine similarity}:}  ConveRT~\citep{henderson2020convert} is a light-weight conversational model pre-trained on the large Reddit conversational corpus~\citep{henderson2019repository}. It provides powerful representations for conversational data and can be used as a response ranker by comparing the cosine similarity between user utterance and multiple responses. In this work, we not only utilize the pre-trained ConveRT for context embedding in \emph{CSC-LSTM}, but also for a target score in \emph{BERT-R}.

The target signal $\tau_{\text{ConveRT-cs}}$ in Section \ref{subsec: BERT re-ranker} means the cosine similarity of ConveRT embedding between user utterance and system response in \emph{BERT-R}.

\textbf{Variation size} measures the variation of the generated system responses, i.e., how many different realisations are generated for one DA on average. The variation size results computed on the full MultiWOZ test data are shown in the left part of the results column ($/$) while the right part shows the variation size computed only over the instances of the test data with DA that appear more than once, i.e., where variation can actually occur.

\section{The Proposed Models and Experiment Details}
\label{sec: the proposed models}
In this section, our proposed models (two contextual generators, one response re-ranker) and corresponding experiment details are introduced.

\begin{table*}
\footnotesize

\begin{center}
\begin{tabular}{l|cccccc}
    \toprule
    context size & \emph{0u0s}  & \emph{1u1s}  & \emph{2u2s}  & \emph{3u3s}  & \emph{4u4s}  & \emph{5u5s} \\
     
   \midrule
   {BLEU-4 (\%)} & 29.59 & 29.76 & 29.68 & 29.50 & \underline{29.46} & \textbf{29.79} \\
   {Meteor (\%)} & \textbf{51.29} & 51.11 & 51.21 & \underline{50.80} & 50.92 & 51.22 \\
   {BERTScore F1 (\%)} & 59.17 & \textbf{59.24} & \underline{59.02} & 59.14 & 59.06 &  59.13\\
   {Variation size} & 2.05 & 2.07 & 2.01 & \underline{1.99} & 2.06 & \textbf{2.11}\\ 
   
   \bottomrule
\end{tabular}
\end{center}

\caption{\label{tab:Results for BLEU-4, Meteor, BERTScore and Variation size of CSC-LSTM model with variant context size.} The results for BLEU, Meteor, BERTScore and Variation size of \emph{CSC-LSTM} model with variant context size (best results are marked with bold font and worst results are marked with underline) show that the best models exist in \emph{0u0s}, \emph{1u1s} and \emph{5u5s}, while the worst models exist in \emph{2u2s}, \emph{3u3s} and \emph{4u4s}.}
\end{table*}

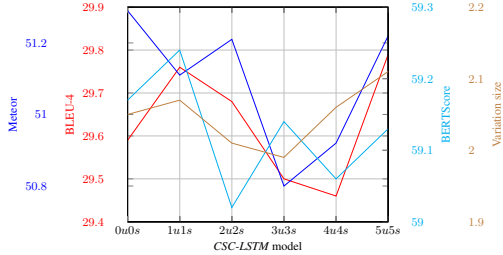
\begin{figure}
\centering
\begin{tikzpicture}[scale=0.5]
\footnotesize
\begin{axis}[
ymin=29.40,
ymax=29.90,
xmin =1,
xmax =6,
grid=both,
axis x line*=bottom,
xlabel = \emph{CSC-LSTM} model,
xlabel near ticks,
xticklabels={$0u0s$, $1u1s$,$2u2s$,$3u3s$,$4u4s$,$5u5s$},
xtick={1,2,3,4,5,6},
ylabel = \textcolor{red}{BLEU-4},
every y tick label/.append style={red},
y tick label style={xshift = -0.5 cm},
]
\addplot[red]
table[x=x,y=y]{x y
1 29.59
2 29.76
3 29.68
4 29.50
5 29.46
6 29.79
};
\end{axis}
\begin{axis}[
ymin=50.70,
ymax=51.30,
xmin =1,
xmax =6,
xticklabels=\empty,
ylabel = \textcolor{blue}{Meteor},
every y tick label/.append style={blue},
y tick label style={xshift = -2 cm},
]
\addplot[blue]
table[x=x,y=y]{x y
1 51.29
2 51.11
3 51.21
4 50.80
5 50.92
6 51.22
};
\end{axis}
\begin{axis}[
ymin=59.00,
ymax=59.30,
xmin =1,
xmax =6,
axis y line*=right,
xticklabels=\empty,
ylabel = \textcolor{cyan}{BERTScore},
every y tick label/.append style={cyan},
y tick label style={xshift = 0.5 cm},
]
\addplot[cyan] 
table[x=x,y=y]{x y
1 59.17
2 59.24
3 59.02
4 59.14
5 59.06
6 59.13
};
\end{axis}
\begin{axis}[
ymin=1.90,
ymax=2.20,
xmin =1,
xmax =6,
axis y line*=right,
xticklabels=\empty,
ylabel = \textcolor{brown}{Variation size},
every y tick label/.append style={brown},
y tick label style={xshift = 2 cm},
]
\addplot[brown]
table[x=x,y=y]{x y
1 2.05
2 2.07
3 2.01
4 1.99
5 2.06
6 2.11
};
\end{axis}
\end{tikzpicture}
\caption{\label{fig:The changeable tendency of all metrics over variable context model.} The curves of BLEU-4, Meteor, BERTScore and Variation size over CSC-LSTM model with different context size (the exact value of metrics please find in Table \ref{tab:Results for BLEU-4, Meteor, BERTScore and Variation size of CSC-LSTM model with variant context size.}) show similar tendency: the both ends of every curve have better performance than the inner part.}
\end{figure}

\subsection{ConveRT initialized SC-LSTM: \emph{CSC-LSTM}}
\label{subsec: convert initialized sc-ltsm}

We train \emph{CSC-LSTM}\footnote{The architecture of \emph{CSC-LSTM} is shown in Appendix.} on the basis of the SC-LSTM~\citep{wen2015semantically}, where a semantic control cell encodes DA into an one-hot embedding to guide the task-oriented generation that is oblivious about any dialogue history. In our proposed \emph{CSC-LSTM}, we apply the pre-trained multi-context ConveRT\footnote{https://github.com/davidalami/ConveRT} for encoding the dialogue history and the contextual embedding is forwarded to initialize the SC-LSTM generator. Before initialization in \emph{CSC-LSTM}, a non-linear transformation\footnote{Using the same hidden size as the dimension of the ConveRT embedding, i.e.,  \emph{CSC-LSTM} (hidden size $512$) without project function $\mathit{d_{0}}$ in Eq.~\ref{equ: non-linear transformation} results in worse performance.} is applied, which is shown in Eq.~\ref{equ: non-linear transformation}, to project the ConveRT embedding into the SC-LSTM decoder space:
\begin{equation}
    \mathit{h_{0} = tanh(\mathbf{W} C_{e} + \mathbf{b})} \; .
\label{equ: non-linear transformation}
\end{equation}
$h_{0}\in\mathbb{R}^{d_{g}}$ is the SC-LSTM decoder initial recurrent state, $\mathit{C_{e}}\in\mathbb{R}^{d_{c}}$ is the ConveRT context embedding and $\mathbf{W}\in\mathbb{R}^{d_{g}\times d_{c}}$ projects the context level embedding into the decoder space. The $\mathbf{W}$ and $\mathbf{b}$ are learnable parameters during the \emph{CSC-LSTM} training.

\paragraph{Experiment details of \emph{CSC-LSTM}: } For \emph{CSC-LSTM}, the SC-LSTM will be used as baseline without additional context information. This means, for each utterance generation, the hidden state is initialized with zeros in SC-LSTM. The MultiWOZ 2.1~\citep{eric2020multiwoz} dataset is used for SC-LSTM and \emph{CSC-LSTM} generation models.

The multi-context ConveRT embedding dimensionality is $512$, hence, the $d_{c}$ in Eq.~\ref{equ: non-linear transformation} is $512$ for \emph{CSC-LSTM} training. In order to ensure a fair comparison, we set the same hyper-parameters for SC-LSTM and \emph{CSC-LSTM}: the hidden size to $300$ (the $d_{g}$ in Eq.~\ref{equ: non-linear transformation}), the learning rate to $5\mathrm{e}{-3}$, the batch size to $128$ and beam search decoding in inference with beam size $10$. Early stopping and cross entropy loss are applied during the SC-LSTM and \emph{CSC-LSTM} training. The responses in SC-LSTM and \emph{CSC-LSTM} are delexicalised text where the slot values are replaced by its corresponding slot tokens.

\paragraph{Context size analysis of \emph{CSC-LSTM}: }
The pre-trained multi-context ConveRT does not only encode the immediate preceding user utterance but in addition a maximum of $10$ previous dialogue sentences, i.e., 5 user utterances and 5 system responses ($5u5s$). To analyse the effect of this context on the performance of the \emph{CSC-LSTM}, multiple models with different context sizes have been trained. And we plot the trend and show exact values of all metric scores in Figure~\ref{fig:The changeable tendency of all metrics over variable context model.} and Table \ref{tab:Results for BLEU-4, Meteor, BERTScore and Variation size of CSC-LSTM model with variant context size.} respectively. The $0u0s$ in Figure~\ref{fig:The changeable tendency of all metrics over variable context model.} and Table \ref{tab:Results for BLEU-4, Meteor, BERTScore and Variation size of CSC-LSTM model with variant context size.} means only immediate preceding user utterance without extra context is taken into account for contextual generator \emph{CSC-LSTM} training.

\subsection{Contextually and Semantically Conditioned GPT: \emph{CSC-GPT}}
\label{subsec:contextually and semantically Condditioned GPT}
In addition to the contextual generator trained from scratch in Section~\ref{subsec: convert initialized sc-ltsm}, we also explore contextual generation adapted from a pre-trained LM model in this section. 

We train \emph{CSC-GPT}\footnote{The architecture of \emph{CSC-GPT} is shown in Figure \ref{fig: architecture of CSC-GPT} in Appendix} on the basis of the pre-trained GPT-2~\citep{radford2019language}. It adopts the generic Transformers~\citep{vaswani2017attention}. \citet{peng2020few} already proposed the SC-GPT model, which was continuously training the GPT-2 on (DA, system response) pairs. However, no context information was taken into account in SC-GPT for dialogue response generation. In our proposed \emph{CSC-GPT}, we leverage the extra user context beyond semantic information to guide the generation process. This means, (user utterance, DA, system response) MultiWOZ triplets are continuously trained on the pre-trained GPT-2 model for contextual generation. Given the extra context size analysis result of \emph{CSC-LSTM} model (shown in \ref{subsec: experiment results of CSC-LSTM}) and GPU memory limitation for training the pre-trained LM, only the immediate preceding user utterance is used as context information in the \emph{CSC-GPT}\footnote{In the Appendix, we also conduct the \emph{CSC-GPT} with different context size: \emph{CSC-GPT (0u0s)} and \emph{CSC-GPT (1u1s)}. The performance comparison in Table \ref{tab:Performance of 0u0s CSC-GPT and 1u1s CSC-GPT.} supports the extra context size analysis in \emph{CSC-LSTM}: longer dialogue context can not linearly improve the generation performance.}.

In this work, we tackle the generation problem using conditional LM. Given the dialogue dataset $\mathcal{D} = \{(\emph{u}_{n}, \emph{d}_{n}, \emph{r}_{n})\}_{n=1}^{N} $ with $N$ samples, the goal is to build a statistical model parameterized by $\theta$ to characterize $p_{\theta}(\emph{r}|\emph{u}, \emph{d})$, which can be written as the product of a series of conditional probabilities.
\begin{equation}
\label{equ: conditional lm}
    p_{\theta}(\emph{r}|\emph{u}, \emph{d}) = \prod_{t=1}^{T} p_{\theta}(r_{t}|r_{<t}, \emph{u}, \emph{d})
\end{equation}
where $r_{<t}$ indicates all tokens before $t$. The $\emph{u}$ represents user utterance; $\emph{d}$ means the system DA and $\emph{r}$ is the system response which includes $(r_{1}, r_{2}, ... r_{t}, ...)$ tokens with length $T$.

\paragraph{Experiment details of \emph{CSC-GPT}: }In order to achieve a robust performance comparison, two datasets, namely MultiWOZ 2.1 and MultiWOZ 2.2, are used in SC-GPT and \emph{CSC-GPT}. During training, the batch size is $16$, the maximal epoch is $10$, the learning rate is $5\mathrm{e}{-5}$ and early stopping is used. During decoding, we use the top-k~\citep{fan2018hierarchical} and nucleus sampling (top-p)~\citep{holtzman2019curious} decoding algorithms with top-k equal to $5$ and top-p equal to $0.9$. This means, the next token distribution is filtered to keep maximal top $5$ tokens with highest probability and the cumulative probability above a $0.9$ threshold. Due to the computational expense of running large SC-GPT and \emph{CSC-GPT} model, only the top $5$ responses are generated.

\subsection{BERT Re-ranker: \emph{BERT-R}}
\label{subsec: BERT re-ranker}
In this paper, we propose a \textbf{BERT}~\citep{devlin2019bert} \textbf{R}e-ranker (\emph{BERT-R}) to select the top generation which is more similar to human sentence and better fits to user context. As the generated responses of the \emph{CSC-LSTM} are delexicalized and have less variability compared with the \emph{CSC-GPT}, we only apply the re-ranker to the \emph{CSC-GPT} model. Adapted from pre-trained BERT, the \emph{BERT-R} is continually trained with task-oriented dialogue data and then fine-tuned on a multiple regression task, where the model learns the relationship between user utterance and system response from the various regression targets defined by multiple metrics scores.

There are two steps in our proposed \emph{BERT-R}: masked LM pre-training and multiple regression fine-tuning. And the BERT-base-uncased\footnote{https://huggingface.co/bert-base-uncased} model with $12$ layers, $768$ hidden units and $12$ heads is used in this work.

\paragraph{Masked LM Pre-training}
The original BERT was pre-trained with the BooksCorpus~\citep{zhu2015aligning} and English Wikipedia. In order to better generalize to task-oriented dialogues, we continually train the BERT model with a dialogue dataset: DSTC8~\citep{rastogi2020towards}, which is a schema-guided dialogue dataset and consists of over $\text{20k}$ annotated multi-domain, task-oriented conversations between a human and a virtual assistant. And \citet{sellam2020bleurt} and \citet{peng2020few} both continually trained with task-specific data based on the pre-trained LMs for better generalisation. Similar to the masked LM training of the original BERT, only $15\%$ tokens are randomly masked for prediction with cross entropy loss.

\paragraph{Multiple Regression Fine-tuning}
The ideal generated response should be close to human communication and relevant to preceding user utterance at the same time. The general idea for \emph{BERT-R} is straightforward: using multiple widely used metric scores to guide the model to learn the relationship of user utterance and system response by a multiple regression task, where the regression targets are those metric scores. Those multiple metric scores define how good the generation is from different perspectives to avoid dominance by one single score. Hence, we employ two $n$-gram matching metrics: BLEU-4 and Meteor score, and a machine learned score: BERTScore, to define how similar system response is with respect to gold reference; and ConveRT cosine similarity to define how contextual system response is with respect to user utterance.

In this work, we continually train \emph{BERT-R} with the guidance of multiple metric scores. Define the user utterance $\emph{u} = (u_{1}, ..., u_{l})$ of length $l$ where each $u_{i}$ is a token and system response $\emph{r} = (r_{1}, ..., r_{m})$ of length $m$. Let $\mathcal{D^\prime} = \{ (\emph{u}_{n}, \emph{r}_{n}, \emph{y}_{n}^{\tau}) \}_{n=1}^{N^\prime}$ be a training dataset of size $N^\prime$, where ${\tau}$ is a target signal. $\emph{y}^{\tau}$ represents all metric scores: $\emph{y}^{\tau} = (\emph{y}^{\tau_{\text{BLEU-4}}},\emph{y}^{\tau_{\text{Meteor}}},\emph{y}^{\tau_{\text{BERT}_{\text{pre}}}},\emph{y}^{\tau_{\text{BERT}_{\text{rec}}}},\emph{y}^{\tau_{\text{BERT}_{\text{f1}}}},\emph{y}^{\tau_{\text{ConveRT-cs}}})$. Hence, the size of $\tau$ is $6$, which means $6$ specific regression layers will be added to the output of [CLS] token in \emph{BERT-R}. $\emph{y}^{\tau}$ will guide \emph{BERT-R} to learn how similar system response $\emph{r}$ is with respect to gold reference and how contextual system response $\emph{r}$ is with respect to user utterance $\emph{u}$ during the fine-tuning. Given the training data, the goal of fine-tuning is to learn a multiple regression function $\emph{f}: (\emph{u}, \emph{r}) \xrightarrow{} \emph{y}^{\tau}$ that predicts different metric scores.

Given the sentence pair $(\emph{u}, \emph{r})$, the pre-trained \emph{BERT-R} returns a sequence of contextualized vectors:
\begin{equation}
    \emph{v}_{\text{[CLS]}}, \emph{v}_{u_{1}}, ..., \emph{v}_{u_{l}}, \emph{v}_{r_{1}}, ..., \emph{v}_{r_{m}} = \emph{BERT-R}(\emph{u}, \emph{r})
\end{equation} 
where $\emph{v}_{\text{[CLS]}}$ is the \emph{BERT-R} output representation for the special [CLS] token, which can be further fine-tuned for classification or regression task. As described by~\citet{devlin2019bert}, we add separate linear layers on top of the [CLS] vector to predict different metric scores:
\begin{equation}
\label{equ: regression function}
    \hat{y}^{\tau} = \emph{f}(\emph{u}, \emph{r}) = \mathbf{W}^{\tau}\emph{v}_{\text{[CLS]}} + \mathbf{b}^{\tau}
\end{equation}
where $\mathbf{W}^{\tau}$ and $\mathbf{b}^{\tau}$ are the weight matrix and bias vector respectively. And we use the Eq.~\ref{equ: multiple regression loss}, the sum of all target-specific regression loss to fine-tune \emph{BERT-R}.
\begin{equation}
\label{equ: multiple regression loss}
    l_{\text{fine-tuning}} = \sum_{\tau} (\frac{1}{N^\prime} \sum_{n=1}^{N^\prime} || \hat{y}_{n}^{\tau} - y_{n}^{\tau} ||^{2})
\end{equation}
After fine-tuning, the \emph{BERT-R} is used to select the top generation with the highest score: the sum of all regression output of \emph{BERT-R}, which is shown in Eq.~\ref{equ: re-ranker score}:
\begin{equation}
\label{equ: re-ranker score}
S_{\emph{BERT-R}} = \sum_{\tau} \emph{BERT-R}(\emph{u}, \emph{r})
\end{equation}


\paragraph{Experiment details of \emph{BERT-R}: }
During masked LM pre-training, batch size is $32$, maximal epoch is $10$ and learning rate is $5\mathrm{e}{-5}$. And early stopping is used to avoid over-fitting on the DSTC8 training dataset. 

For the fine-tuning of \emph{BERT-R}, we generate the top $5$ responses for MultiWOZ data with \emph{CSC-GPT} model firstly. And in order to clean the system generated responses, we have the following procedures:
\begin{itemize}
  \item [1)]
The duplicated system responses are removed.
  \item [2)]
The last turn of every dialogue is removed, where there are always ``thank you'' and ``good bye'', kind of non-informative sentences.
  \item [3)]
In order to let the network glance what the human communication looks like, we add the gold reference for the user utterance in the training dataset. In addition, we need to remove the system responses which are same as the gold reference beforehand, to comply with the rule 1).
\end{itemize}
After finishing the above process, we compute the target score respectively for the label of different regression layers. During the training of the multiple regression task, the batch size is set to $32$, learning rate $1\mathrm{e}{-5}$, and early stopping is used to save the best BERT re-ranker.

\begin{table*}[h!]
\setlength{\tabcolsep}{4pt}
\footnotesize
\begin{center}
\begin{tabular}{lcccccccc}
    \toprule
    & \multicolumn{4}{c}{MultiWOZ 2.1} & \multicolumn{4}{c}{MultiWOZ 2.2} \\
    \cmidrule(lr{.75em}){2-5}  \cmidrule(lr{.75em}){6-9}
    
    & \makecell{BLEU-4 \\ (\%)} & \makecell{Meteor \\(\%)} & \makecell{BERTScore \\ F1 (\%)} & \makecell{Variation \\ size} &
    \makecell{BLEU-4 \\ (\%)} & \makecell{Meteor \\(\%)} & \makecell{BERTScore \\ F1 (\%)} & \makecell{Variation \\ size}\\ 
    
    \midrule
   SC-LSTM~\cite{wen2015semantically} & 28.76 & 49.93  & 58.71 & 1.00/1.00
   & - & -  & - & - \\

   \emph{CSC-LSTM} ($5u5s$) & \textbf{29.79} & \textbf{51.22} &  \textbf{59.13} & \textbf{1.35/2.11} & 
   - & - &  - & - \\

    \cmidrule{2-9}
   SC-GPT~\cite{peng2020few} & 28.95 & 50.22  & 91.96 & 2.27/6.56
   & 28.53 & 49.80  & 91.95 & 2.78/6.70 \\

   \emph{CSC-GPT} & {29.91} & {51.34} &  {92.08} & \textbf{2.29/6.66} & 
   {29.41} & {51.10} &  {92.08} & \textbf{2.81/6.82} \\

   \emph{BERT-R} & \textbf{32.37} & \textbf{54.01} &  \textbf{92.40} & 2.22/6.34
   & \textbf{31.68} & \textbf{53.65} &  \textbf{92.39} & \textbf{2.81/6.82}\\

   \bottomrule
\end{tabular}
\end{center}
\caption{\label{tab:Performance comparison on automatic metrics of all models with top one generation} The results for BLEU, Meteor, BERTScore and variation size of top one generation in all models demonstrate that our proposed contextual models in: \emph{CSC-LSTM} against SC-LSTM and \emph{CSC-GPT} against SC-GPT, both outperform the corresponding baseline. Meanwhile, the proposed re-ranker \emph{BERT-R} also highly improve the metric scores compared with all other models.}
\end{table*}

\section{The Experiment Results}
\label{sec: the experiment results}
The experimental results of all models introduced in Section \ref{sec: the proposed models} are presented in this section. To ensure a consistent performance comparison, we compute the metric scores based on the top one generation of all models. Table \ref{tab:Performance comparison on automatic metrics of all models with top one generation} shows the results of all automatic metrics.

Furthermore, a \textbf{human evaluation} has been conducted. We randomly sampled $100$ dialogues and their corresponding top one generations from our proposed models as well as the baselines. We recruited three annotators with relevant background in SDS to evaluate the responses generated by different models. Each rater was presented the complete preceding dialogue and asked to rate if ``The highlighted system response could plausibly have been produced by a human'' (natural) and if ``The highlighted system response fits well to the previous dialog'' (contextual). Each metric is rated on a 5-point Likert scale, where $1$ is ``not agree at all'', $5$ is ``fully agree''. In order to guarantee the strictness of human evaluation, the human judges have no information about the origin of the utterances, i.e., which model generated the utterance. Table \ref{tab: human evaluation results.} shows the human evaluation results.

\begin{table}[h]
\footnotesize
\begin{center}
\begin{tabular}{lcc}
    \toprule
    
     & Contextual & Natural \\ 
     
   \midrule
   SC-LSTM~\cite{wen2015semantically} & 3.96 & 4.04 \\
   \emph{CSC-LSTM} ($5u5s$) & \textbf{4.21$^{*}$} & \textbf{4.16$^{*}$} \\
   \cmidrule{2-3}
   SC-GPT~\cite{peng2020few} & 4.00 & 4.14\\
   \emph{CSC-GPT} & \textbf{4.25$^{+}$} & \textbf{4.27$^{+}$} \\
   \emph{BERT-R} & 4.18 & 4.26 \\
   
   %

   \bottomrule
\end{tabular}
\end{center}
\caption{\label{tab: human evaluation results.} The results of human evaluation on natural and contextual score of all models. ($*$: $p$-value $<$ 0.01, comparison with SC-LSTM baseline; $+$: $p$-value $<$ 0.1, comparison with SC-GPT baseline) show the superiority of our proposed contextual models.}
\end{table}

\subsection{Experiment results of \emph{CSC-LSTM}}
\label{subsec: experiment results of CSC-LSTM}
The automatic metric scores comparing SC-LSTM and \emph{CSC-LSTM} in Table~\ref{tab:Performance comparison on automatic metrics of all models with top one generation} show that \emph{CSC-LSTM} has the overall better performance in BLEU-4, Meteor, BERTScore and variation size compared to the baseline. The variation size results show \emph{CSC-LSTM} can generate more variant responses per DA, which may indicate a more contextual fitting response, while the SC-LSTM only generate the same utterances each time. The performance comparison between SC-LSTM and \emph{CSC-LSTM} in Table~\ref{tab:Performance comparison on automatic metrics of all models with top one generation} support our initial assumption that context helps to generate good system utterances. Especially the increase in variation size is of importance as it indicates that the resulting utterances of \emph{CSC-LSTM} indeed be different for different contexts.

This has been validated by the human evaluation of SC-LSTM and \emph{CSC-LSTM} in Table \ref{tab: human evaluation results.}. It shows that the variation introduced by \emph{CSC-LSTM} actually results in utterances that fit significantly better to the preceding dialogue and are perceived as significantly more natural. Overall, the performance comparison between SC-LSTM and \emph{CSC-LSTM} on automatic metrics and human evaluation demonstrate the dialogue history contributes to contextual and variant responses.

The context size analysis of \emph{CSC-LSTM} demonstrate that the automatic metric scores are influenced by the length of the context. All metrics show similar curves over the different contextual model in Figure~\ref{fig:The changeable tendency of all metrics over variable context model.}. The both ends of the curves have better performance than the inner part. The Table \ref{tab:Results for BLEU-4, Meteor, BERTScore and Variation size of CSC-LSTM model with variant context size.} shows the best BLEU-4 and variation size are both achieved for context sizes of $5u5s$; while $0u0s$ has the best Meteor and $1u1s$ has the best BERTScore. Both show that the contextual models $0u0s$, $1u1s$ and $5u5s$ generally outperform the $2u2s$, $3u3s$ and $4u4s$. Hence, the investigation of the impact of context size for \emph{CSC-LSTM} generation indicates that longer context does not linearly result in better performance, which is further confirmed in the performance comparison between \emph{CSC-GPT}($0u0s$) and \emph{CSC-GPT}($1u1s$) shown in Table \ref{tab:Performance of 0u0s CSC-GPT and 1u1s CSC-GPT.} in Appendix. Evidently, all contextual models achieve better performance than the baseline (SC-LSTM in Table \ref{tab:Performance comparison on automatic metrics of all models with top one generation}). We therefore conclude that the immediate preceding user utterance yields the indispensable information for contextual generation. With limited memory, using only immediate preceding user utterance without extra context can be regarded as a balanced option that we directly apply for training the \emph{CSC-GPT}.

\subsection{Experiment results of \emph{CSC-GPT}}
\label{subsec: experiment results of CSC-GPT}
All scores between SC-GPT and \emph{CSC-GPT} in Table~\ref{tab:Performance comparison on automatic metrics of all models with top one generation} demonstrate that the \emph{CSC-GPT} is superior to the baseline SC-GPT for both datasets, MultiWOZ 2.1 and MultiWOZ 2.2. Our assumption is again confirmed in GPT-based generation model: dialogue history contributes to contextual and variant response, even though only one preceding user utterance is taken into account.

When comparing the human evaluation results of SC-GPT and \emph{CSC-GPT} in Table \ref{tab: human evaluation results.}, the assumption is further supported: adding context to the generation process results in more natural and contextual responses. The \emph{CSC-GPT} generator even achieves the best rating both in terms of natural and contextual score compared to all other models. This means that enhancing an already powerful pre-trained model with context is essential for its application within dialogue systems.

\subsection{Experiment results of \emph{BERT-R}}
\label{subsec: BERT-R Results}
Our proposed \emph{BERT-R} selects the top one response from $5$ \emph{CSC-GPT} generations with the highest score in Eq. \ref{equ: re-ranker score}. The metric scores of \emph{CSC-GPT} and \emph{BERT-R} in Table \ref{tab:Performance comparison on automatic metrics of all models with top one generation} show that the selected generation by \emph{BERT-R} has a significant improvement on BLEU-4, Meteor and BERTScore\footnote{also improvement on ConveRT cosine similarity, which is shown in the Appendix.}, with a little loss on variation size compared to the top one generation in \emph{CSC-GPT}. This is in line with human evaluation results where the \emph{CSC-GPT} achieves slightly better scores both in naturalness and contextualness than \emph{BERT-R}, even though \emph{BERT-R} clearly shows better results compared to the SC-GPT baseline.

\section{Conclusion and Future Work}
\label{sec:Conclusion and Future Work}
In this paper, we propose two contextual generation models: \emph{CSC-LSTM} trained from scratch and \emph{CSC-GPT} adapted from pre-trained GPT-2. Both integrate dialogue context information into NLG for generating more variant and contextual response in task-oriented dialogue systems. 

In the experiment of \emph{CSC-LSTM} against SC-LSTM and \emph{CSC-GPT} against SC-GPT, our proposed contextual models both improve the generation performance in automatic metrics, thus showing that \emph{CSC-LSTM} and \emph{CSC-GPT} are able to capture better the contextual needs resulting in a higher similarity to the data. This is further underpinned by the number of variations. More variant responses are generated per DA in \emph{CSC-LSTM}, while the SC-LSTM only generates the same utterances each time. Furthermore, the variation size of GPT-based generators is higher than SC-LSTM based models. The possible reason is the pre-trained GPT-2 contributes to more diverse responses by default. The human evaluation results in Table \ref{tab: human evaluation results.} not only demonstrate the contextual model can generate more contextual and natural response compared with their baseline respectively, but also show GPT-2 contextual model \emph{CSC-GPT} is superior than \emph{CSC-LSTM}.

An investigation of the impact of context size for dialogue response generation in \emph{CSC-LSTM} indicates that longer context does not automatically result in better performance. However, all variant \emph{CSC-LSTM} models have better performance than baseline, which means the immediate preceding user utterance contains the most contextual information for generation. This is also verified in GPT-2 contextual generators, even only immediate preceding user utterance is taken into account, the \emph{CSC-GPT} model outperforms SC-GPT both on automatic metrics and human evaluation.

In addition to the above mentioned two contextual models, we also present a re-ranker for \emph{CSC-GPT} contextual model. Adapted from pre-trained BERT, the \emph{BERT-R} continually train on multi-domain dialogues and fine-tune on a multiple regression task to learn the relationship between user context and system response by the metric guidance of BLEU-4, Meteor, BERTScore and ConveRT cosine similarity. Finally, the top one generation selected by \emph{BERT-R} has significant superiority in BLEU-4, Meteor, BERTScore and ConveRT cosine similarity compared with top one generation in \emph{CSC-GPT}. This means, that our proposed \emph{BERT-R} works from the guidance of metric scores and can choose the generation with highest score. However, \emph{CSC-GPT} slightly outperforms \emph{BERT-R} in variation size and human evaluation. The possible reason is that the existing automatic metrics still have bias with human judgments \citep{chaganty2018price}.

In the future, we will further explore the performance of \emph{BERT-R} with the guidance of other automatic metrics, which have higher correlation with human judgements. Furthermore, there seems to be a link between the variation size metric and the human evaluation scores, which will also be part of future work.


\bibliographystyle{acl_natbib}
\bibliography{anthology,acl2020}

\appendix

\section{Appendices}
\label{sec:appendix}

\begin{figure*}
\centering
\begin{tikzpicture}[scale=1]

\draw[thick,rounded corners]   (-3,-1) rectangle (9,1);
\node at (-2,0.5)[align=center]  {GPT-2};
\draw [decorate,decoration={brace,amplitude=5pt,raise=5pt,mirror}] (-3,-1) -- (1,-1) node [black,midway,yshift=-0.6cm] {$u_{n}$};
\draw [decorate,decoration={brace,amplitude=5pt,raise=5pt,mirror}] (1,-1) -- (5,-1) node [black,midway,yshift=-0.6cm] {$d_{n}$};
\draw [decorate,decoration={brace,amplitude=5pt,raise=5pt,mirror}] (5,-1) -- (9,-1) node [black,midway,yshift=-0.6cm] {$r_{n}$};
\draw [decorate,decoration={brace,amplitude=5pt,raise=5pt}] (5,1) -- (9,1) node [black,midway,yshift=0.6cm] {$r_{n}$};

\draw[->,thick, dash dot] (-1.5,-0.8) -- (7,0.8);
\draw[->,thick, dash dot] (0,-0.8) -- (7,0.8);

\draw[->,thick, dash dot] (2.5,-0.8) -- (7,0.8);
\draw[->,thick, dash dot] (4,-0.8) -- (7,0.8);

\draw[->,thick, dash dot] (7,-0.8) -- (7,0.8);

\draw[thick, dotted] (-2.5, 0) -- (1, 0);
\draw[thick, dotted] (6, 0) -- (6.7, 0);
\draw[thick, dotted] (7.3, 0) -- (8.7, 0);

\node at (3,-2.5)[align=center]  {
$u_{n}$ is the immediate preceding user utterance, like ``Does it provide free parking?''\\
$d_{n}$ is the DA, like ``hotel \{inform (internet=yes; parking=yes)\}''\\
$r_{n}$ is the gold response, like ``Yes, free parking and free wifi!''};
\end{tikzpicture}
\caption{\label{fig: architecture of CSC-GPT} The architecture of \emph{CSC-GPT}. The immediate preceding user utterance and MR together guide the contextual response generation.}
\end{figure*}
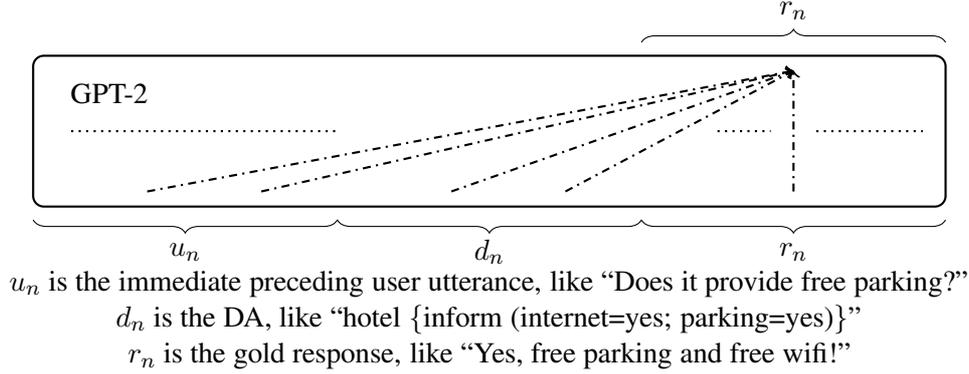

In order to better understand what our proposed architectures look like, please find the following Figure \ref{fig: architecture of CSC-LSTM} and Figure \ref{fig: architecture of CSC-GPT}. The model \emph{CSC-GPT} generates contextual response with the guidance of immediate preceding user utterance and DA. In the \emph{CSC-LSTM}, the SC-LSTM was proposed in \citet{wen2015semantically} and an additional cell was introduced into the LSTM cell to gate the DA information. The original LSTM cell follows:
\begin{equation}
\begin{aligned}
    i_{t} &= \sigma(\mathbf{W}_{wi}w_{t}+\mathbf{W}_{hi}h_{t-1}) \\
    f_{t} &= \sigma(\mathbf{W}_{wf}w_{t}+\mathbf{W}_{hf}h_{t-1}) \\
    o_{t} &= \sigma(\mathbf{W}_{wo}w_{t}+\mathbf{W}_{ho}h_{t-1}) \\
    \hat{c}_{t} &= \tanh(\mathbf{W}_{wc}w_{t}+\mathbf{W}_{hc}h_{t-1}) \\
    c_{t} &= f_{t} \odot c_{t-1} + i_{t} \odot \hat{c}_{t} \\
    h_{t} &= o_{t} \odot \tanh(c_{t})
\label{equ: original LSTM}
\end{aligned}   
\end{equation}
In SC-LSTM, the $d_{0}$ starts from an one-hot vector, at each time step the DA cell decides what information should be retained for future time steps and discards the others. Like:
\begin{equation}
\begin{aligned}
    r_{t} &= \sigma(\mathbf{W}_{wr}w_{t}+ \alpha (\mathbf{W}_{hr}h_{t-1})) \\
    d_{t} &= r_{t} \cdot d_{t-1}
\label{equ: DA in SC-LSTM}
\end{aligned}   
\end{equation}
Then, the value cell in Eq. \ref{equ: original LSTM} also depends on the DA,
\begin{equation}
    c_{t} = f_{t} \odot c_{t-1} + i_{t} \odot \hat{c}_{t} + \tanh(\mathbf{W}_{dc}d_{t})
\end{equation}
Finally, the hidden state is further updated by new value cell. In our proposed \emph{CSC-LSTM}, not only DA cell is added, but the SC-LSTM cell is initialized by contextual ConveRT embedding.

\begin{figure*}
\centering
\begin{tikzpicture}[scale=1]

\draw[thick, dotted] (-6.5, 0) -- (-5.5, 0);

\draw[->,thick] (-5,-0.5) -- (-4,-0.5);
\node at (-4.5,0.7)[align=center]  {$h_{t-1}$};
\draw[->,thick] (-5,0.5) -- (-4,0.5);
\node at (-4.5,-0.3)[align=center]  {$d_{t-1}$};

\draw[thick,rounded corners]   (-4,-1) rectangle (-1,1);
\node at (-2.5,0)[align=center]  {SC-LSTM};

\draw[->,thick] (-2.5,-1.5) -- (-2.5,-1);
\node at (-2.2,-1.4)[align=center]  {$w_{t}$};

\draw[->,thick] (-1,-0.5) -- (0,-0.5);
\node at (-0.5,0.7)[align=center]  {$h_{t}$};
\draw[->,thick] (-1,0.5) -- (0,0.5);
\node at (-0.5,-0.3)[align=center]  {$d_{t}$};

\draw[thick,rounded corners]   (0,-1) rectangle (3,1);
\node at (1.5,0)[align=center]  {SC-LSTM};

\draw[->,thick] (1.5,-1.5) -- (1.5,-1);
\node at (2.0,-1.4)[align=center]  {$w_{t+1}$};

\draw[->,thick] (3,-0.5) -- (4,-0.5);
\node at (3.5,0.7)[align=center]  {$h_{t+1}$};
\draw[->,thick] (3,0.5) -- (4,0.5);
\node at (3.5,-0.3)[align=center]  {$d_{t+1}$};

\draw[thick, dotted] (4.5, 0) -- (5.5, 0);

\node at (0,-2.5)[align=center]  {
$h_{0}$ refers to Eq.~\ref{equ: non-linear transformation}, dialogue context is encoded by ConveRT and projected to decoder space with the Eq.~\ref{equ: non-linear transformation}\\
$d_{0}$ is DA one-hot embedding, like (0, 0, ..., 1, ...) represents ``hotel\{ request(area=?)\}''\\
$w_{t}$ is the input token at time slot $t$};

\end{tikzpicture}
\caption{\label{fig: architecture of CSC-LSTM} The architecture of \emph{CSC-LSTM}.}
\end{figure*}

Given the pre-trained model can handle the longer dialogue context for generation, so we also trained \emph{CSC-GPT} contextual model with one more turn context besides the immediate preceding user utterance, i.e. $1u1s$. And the results comparison between \emph{CSC-GPT (0u0s)} and \emph{CSC-GPT (1u1s)} is shown in Table \ref{tab:Performance of 0u0s CSC-GPT and 1u1s CSC-GPT.}. We can know that the BLEU-4 and Meteor of \emph{CSC-GPT (0u0s)} outperforms \emph{CSC-GPT (1u1s)}, meanwhile, the BERTScore and variation size have no big difference. This further demonstrate that longer context can not directly result in better performance. The maximal length of input in \emph{CSC-GPT (0u0s)} is $120$, however, \emph{CSC-GPT (1u1s)} is $190$ with more GPU memory. Hence, we recommend only immediate preceding user utterance is taken into account for contextual generation in task-oriented dialogue system.

\begin{table*}[h!]
\setlength{\tabcolsep}{4pt}
\footnotesize

\begin{center}
\begin{tabular}{lcccccccc}
    \toprule
    & \multicolumn{4}{c}{MultiWOZ 2.1} & \multicolumn{4}{c}{MultiWOZ 2.2} \\
    \cmidrule(lr{.75em}){2-5}  \cmidrule(lr{.75em}){6-9}
    
    & \makecell{BLEU-4 \\ (\%)} & \makecell{Meteor \\(\%)} & \makecell{BERTScore \\ F1 (\%)}  & \makecell{Variation \\ size} &
    \makecell{BLEU-4 \\ (\%)} & \makecell{Meteor \\(\%)} & \makecell{BERTScore \\ F1 (\%)} & \makecell{Variation \\ size}\\ 
    
   \midrule
   \emph{CSC-GPT (0u0s)} & \textbf{29.91} & \textbf{51.34} &  \textbf{92.08} & \textbf{2.29}/6.66 & \textbf{29.41} & \textbf{51.10} &  \textbf{92.08} & \textbf{2.81/6.82} \\

   \emph{CSC-GPT (1u1s)} & 29.70 & 51.02 &  \textbf{92.08} & \textbf{2.29/6.67}
   & 28.80 & 50.32 &  91.94 & \textbf{2.81}/6.81\\

   \bottomrule
\end{tabular}
\end{center}

\caption{\label{tab:Performance of 0u0s CSC-GPT and 1u1s CSC-GPT.} The results for BLEU, Meteor, BERTScore and Variation size of top one \emph{CSC-GPT (0u0s)} generation and \emph{CSC-GPT (1u1s)} show that \emph{CSC-GPT (0u0s)} outperforms \emph{CSC-GPT (1u1s)} and support the context size analysis of \emph{CSC-LSTM}: the longer context in contextual model can not linearly result in better performance.}
\end{table*}

The \emph{BERT-R} is trained with multiple metric scores: BLEU-4, Meteor, BERTScore and ConveRT cosine similarity. In order to make consistent comparison of all models, we don't show the ConveRT cosine similarity of \emph{BERT-R} in the main paper. The Table \ref{tab:Performance of reslected generations by BERT-R and top one CSC-GPT.} shows that all scores get improvement.

\begin{table*}[h!]
\setlength{\tabcolsep}{4pt}
\footnotesize

\begin{center}
\begin{tabular}{lcccccccc}
    \toprule
    & \multicolumn{4}{c}{MultiWOZ 2.1} & \multicolumn{4}{c}{MultiWOZ 2.2} \\
    \cmidrule(lr{.75em}){2-5}  \cmidrule(lr{.75em}){6-9}
    
    & \makecell{BLEU-4 \\ (\%)} & \makecell{Meteor \\(\%)} & \makecell{BERTScore \\ F1 (\%)}  & \makecell{ConveRT \\ cs} &
    \makecell{BLEU-4 \\ (\%)} & \makecell{Meteor \\(\%)} & \makecell{BERTScore \\ F1 (\%)} & \makecell{ConveRT \\ cs}\\ 
    
   \midrule
   \emph{CSC-GPT} & 29.91 & 51.34 &  92.08  & 33.41 & 29.41 & 51.10 & 92.08 & 33.81\\
   
   \emph{BERT-R} & \textbf{32.37} & \textbf{54.01} &  \textbf{92.40} & \textbf{34.70}
   & \textbf{31.68} & \textbf{53.65} &  \textbf{92.39} & \textbf{35.00}\\

   \bottomrule
\end{tabular}
\end{center}
\caption{\label{tab:Performance of reslected generations by BERT-R and top one CSC-GPT.} The results for BLEU, Meteor, BERTScore and ConveRT cosine similarity of top one \emph{CSC-GPT} generation and the selected one by \emph{BERT-R} demonstrate that all target metrics used in \emph{BERT-R} have a significant improvement through re-ranking.}
\end{table*}

There are several use cases in Table \ref{tab:System response of different models given preceding dialogue context.}. And by system response comparison of SC-LSTM agaist \emph{CSC-LSTM} and SC-GPT agaist \emph{CSC-GPT} given preceding dialogue context, the Table \ref{tab:System response of different models given preceding dialogue context.} shows the importance of dialogue context for natural and contextual response.

\begin{table*}

\begin{center}
\begin{tabular}{cc}
    \toprule
    context & \makecell[l]{\textbf{User:} Hi, I'm looking for a hotel called aylesbray lodge guest house. Do you \\ have anything with that name?} \\
    \midrule
    system response &   \makecell[l]{SC-LSTM: It is a moderately priced hotel in the south. Would you like me \\ to book it for you?
                \\ \emph{CSC-LSTM}: \underline{Yes,} it is a hotel in the south in the moderately priced price range. \\ Would you like to book a room?
                \\ SC-GPT: It is a moderately priced hotel in the south area. Would you like me \\ to book it for you?
                \\ \emph{CSC-GPT}: \underline{I do,} it is a moderately priced hotel in the south. Would you like \\ to make a booking?}\\
    \midrule
    \midrule
    context & \makecell[l]{\textbf{User:} I'm looking for a nightclub to go to. Are there any in the centre of town?
                    \\ \textbf{System:} There are 5 nightclubs in the centre. Ballare is one of them. Would that \\ work for you?
                    \\ \textbf{User:} Is there an entrance fee for Ballare?} \\
    \midrule
    system response &  \makecell[l]{SC-LSTM: The entrance fee is 5 pounds.
                \\ \emph{CSC-LSTM} \underline{Yes,} it is 5 pounds .
                \\ SC-GPT: The entrance fee is 5 pounds.
                \\ \emph{CSC-GPT}: \underline{Yes,} it is 5 pounds \underline{to get in.}}\\
    
    \midrule
    \midrule
    context & \makecell[l]{\textbf{User:} Hi, I am trying to plan my trip and could use some help with a particular restaurant.} \\
    \midrule
    system response &  \makecell[l]{SC-LSTM: Do you have a preference for                    area or price range?
                \\ \emph{CSC-LSTM:} Do you have a certain area or price range \underline{in mind?}
                \\ SC-GPT: \underline{Sure,} what kind of food are you looking for and what area and price range?
                \\ \emph{CSC-GPT}: \underline{I would love to assist you!} What type of food are you looking for and \\ what price range?}\\
    
    \midrule
    \midrule
    context & \makecell[l]{\textbf{User:} Can you also help me find a train             to Bishops Stortford? 
                \\ \textbf{System:} There are more than 70 trains to Bishops Stratford. Could you tell me \\ what day you would like to leave, where you would like to leave from, and the times?
                \\ \textbf{User:} I am leaving Cambridge on Sunday, can you please tell me the travel time on that ride?
                \\ \textbf{System:} Sure, it is a 38 minute trip.  Would you like a ticket?
                \\ \textbf{User:} I need to leave after 14:45 also.} \\
    \midrule
    system response &  \makecell[l]{SC-LSTM: The tr3600 leaves Cambridge at                 15:29.
                \\ \emph{CSC-LSTM:} Tr3600 leaves Cambridge at 15:29. \underline{Would that works for you?}
                \\ SC-GPT: Tr3600 leaves Cambridge at 15:29.
                \\ \emph{CSC-GPT}: \underline{How about tr3600?} It leaves Cambridge at 15:29.}\\
    
   \bottomrule
\end{tabular}
\end{center}
\caption{\label{tab:System response of different models given preceding dialogue context.} System response of different models given preceding dialogue context show that the contextual models: \emph{CSC-LSTM} and \emph{CSC-GPT}, can generate more natural responses that fit better to dialogue history.}
\end{table*}

\end{document}